%% file: main.tex
\documentclass[conference]{IEEEtran}
\IEEEoverridecommandlockouts
\usepackage{cite}
\usepackage{amsmath,amssymb,amsfonts}
\usepackage{algorithmic}
\usepackage{graphicx}
\usepackage{textcomp}

\usepackage[table]{xcolor}
\usepackage{multirow}
\usepackage{booktabs}
\usepackage{vcell}
\usepackage{subcaption}
\usepackage[a-1b]{pdfx}

\newcommand{\ours}{FedCTTA}
\newcommand*\gcell{\cellcolor[gray]{0.9}}
\newcommand*\pcm{\printcellmiddle}
\newcommand*\pcb{\printcellbottom}
\newcommand{\rot}[1]{\vcell{\rotatebox{70}{#1}}}
\newcommand*\rotv{\rotatebox{90}}
\newcommand{\brule}{\noalign{\vskip -1.5pt}\bottomrule}
\newcommand{\mrule}{\noalign{\vskip -1.5pt}\midrule\noalign{\vskip -2.5pt}}
\newcommand{\dmrule}{\noalign{\vskip -1.5pt}\midrule\midrule\noalign{\vskip -2.5pt}}
\newcommand{\cmrule}[1]{\noalign{\vskip -1.5pt}\cmidrule{#1}\noalign{\vskip -2.5pt}}

\newcommand{\trule}{\toprule\noalign{\vskip -2.5pt}}

\begin{document}

\title{FedCTTA: A Collaborative Approach to Continual Test-Time Adaptation in Federated Learning}

\author{
    \IEEEauthorblockN{
        Rakibul Hasan Rajib\IEEEauthorrefmark{1},
        Md Akil Raihan Iftee\IEEEauthorrefmark{1},
        Mir Sazzat Hossain\IEEEauthorrefmark{1}, 
        A. K. M. Mahbubur Rahman\IEEEauthorrefmark{1},\\
        Sajib Mistry\IEEEauthorrefmark{2},
        M Ashraful Amin\IEEEauthorrefmark{1} and
        Amin Ahsan Ali\IEEEauthorrefmark{1}
    }
    \IEEEauthorblockA{
        \IEEEauthorrefmark{1}Center for Computational \& Data Sciences, Independent University, Bangladesh
    }
    \IEEEauthorblockA{
        \IEEEauthorrefmark{2}Curtin University
    }
    \IEEEauthorblockA{
        {\tt\small rakibul@iub.edu.bd, iftee1807002@gmail.com, \{sazzat, akmmrahman\}@iub.edu.bd,} \\
        {\tt\small Sajib.Mistry@curtin.edu.au, \{aminmdashraful, aminali\}@iub.edu.bd}
        }
}

\maketitle

\input{sections/0_abstract}
\input{sections/1_intro}
\input{sections/2_related_works}
\input{sections/3_methodology}
\input{sections/4_experimental_results}
\input{sections/5_ablation_study}
\input{sections/6_conclusion}

\bibliographystyle{IEEEtran}
\bibliography{main}

\end{document}

%% file: sections/0_abstract.tex
\begin{abstract}
Federated Learning (FL) enables collaborative model training across distributed clients without sharing raw data, making it ideal for privacy-sensitive applications. However, FL models often suffer performance degradation due to distribution shifts between training and deployment. Test-Time Adaptation (TTA) offers a promising solution by allowing models to adapt using only test samples. However, existing TTA methods in FL face challenges such as computational overhead, privacy risks from feature sharing, and scalability concerns due to memory constraints. To address these limitations, we propose Federated Continual Test-Time Adaptation (FedCTTA), a privacy-preserving and computationally efficient framework for federated adaptation. Unlike prior methods that rely on sharing local feature statistics, FedCTTA avoids direct feature exchange by leveraging similarity-aware aggregation based on model output distributions over randomly generated noise samples. This approach ensures adaptive knowledge sharing while preserving data privacy. Furthermore, FedCTTA minimizes the entropy at each client for continual adaptation, enhancing the model's confidence in evolving target distributions. Our method eliminates the need for server-side training during adaptation and maintains a constant memory footprint, making it scalable even as the number of clients or training rounds increases. Extensive experiments show that FedCTTA surpasses existing methods across diverse temporal and spatial heterogeneity scenarios. 
\end{abstract}


%% file: sections/1_intro.tex
\section{Introduction}
\label{sec:introduction}

\begin{figure*}[!ht]
    \centering
    \includegraphics[width=0.8\linewidth, keepaspectratio]{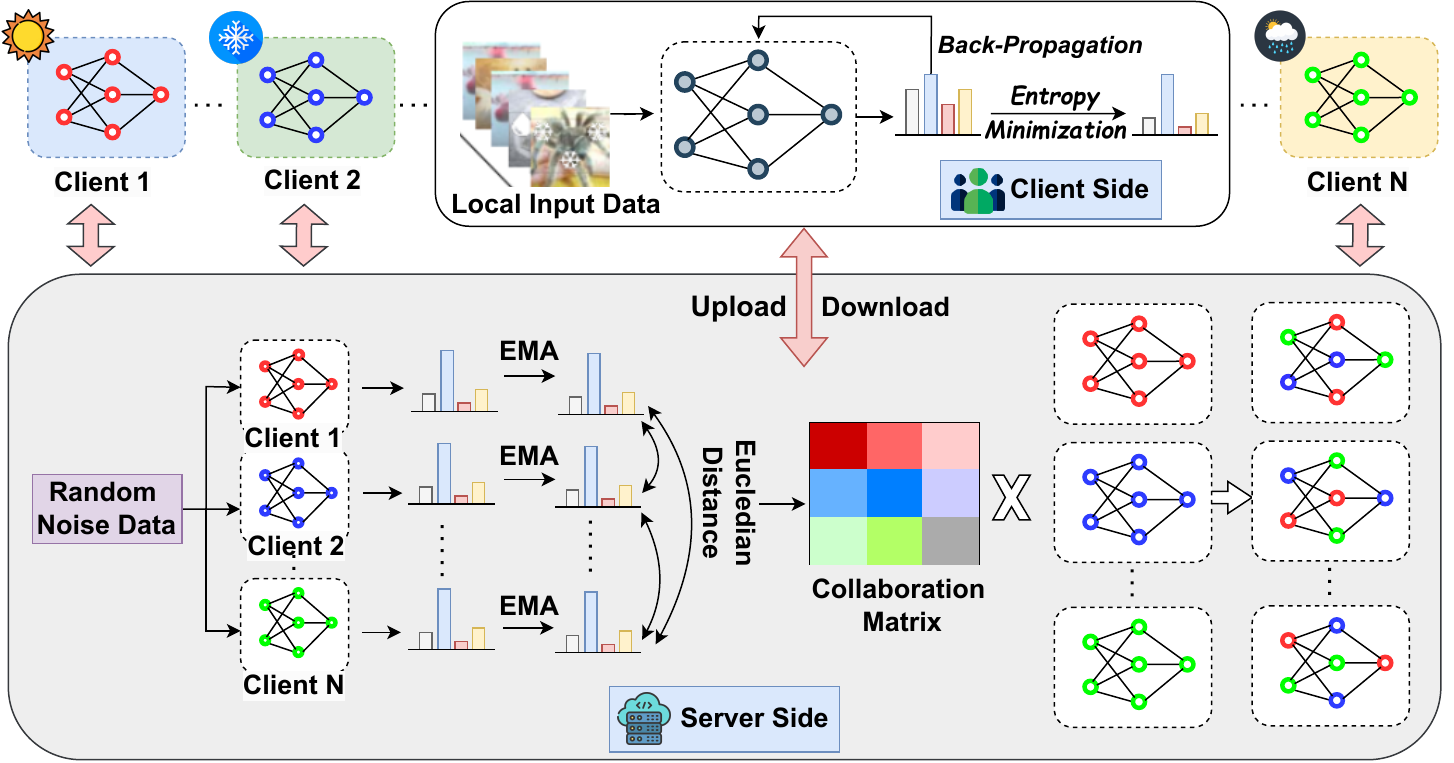}
    \caption{Illustration of the FedCTTA framework. The server aggregates models received from all the clients based on functional similarity, then distributes the personalized aggregated models back to the clients for the next round. This process continues iteratively to adapt the models collaboratively across all clients. The figure depicts both client-to-server communication (model updates) and server-to-client communication (aggregated model distribution).}
    \label{fig:model}
    \vspace{-5mm}
    
\end{figure*}

\textbf{Test-Time Adaptation (TTA)} is revolutionizing deep learning by enabling models to adapt dynamically to unseen data distributions during deployment. Traditional machine learning models often degrade in performance when a distribution shift occurs between training and testing data. For instance, a medical imaging system trained on high-resolution hospital scans may struggle to interpret low-quality scans from rural clinics due to differences in imaging equipment. Existing solutions such as Domain Generalization (DG) \cite{hendrycks2021, hendrycks2020, tremblay2018} and Domain Adaptation (DA) \cite{wang2018, tzeng2017} attempt to mitigate this issue by either training on diverse domains or adapting from a source domain. However, DG requires sufficient domain diversity, and DA depends on access to source data, which may be impractical due to privacy constraints. TTA overcomes these constraints by allowing models to self-adapt using only incoming test samples, eliminating the need for retraining or access to original training data. Recent advancements in TTA \cite{wang2022, bao2024, marsden2023} leverage techniques such as entropy minimization, self-supervised learning, and feature alignment, ensuring robust model performance in dynamic and privacy-sensitive environments.

Data privacy concerns, driven by regulations like GDPR \cite{gdpr2016}, challenge traditional machine learning, which relies on centralized data processing. Federated Learning (FL) \cite{mcmahan2017} addresses this by enabling collaborative model training across decentralized clients without sharing raw data, making it ideal for privacy-sensitive domains like healthcare and finance. However, \textit{TTA in FL remains challenging due to heterogeneous and evolving data distributions across clients.} For instance, in federated healthcare, hospitals generate non-IID data due to varying patient demographics, equipment, and practices, causing models to struggle when deployed in settings with unseen data distributions.

\textit{Local adaptation}, where each client fine-tunes the model using its own test data, fails to leverage broader shifts useful for generalization. \textit{Collaborative adaptation}, where clients share insights without raw data, could improve performance but faces challenges: (1) \textbf{privacy risks} from feature or gradient sharing, (2) \textbf{model misalignment} due to distinct distribution shifts, and (3) \textbf{scalability} issues for resource-constrained clients. Our objective is to \textit{develop privacy-preserving and efficient TTA framework} that enables decentralized models to adapt to evolving conditions while maintaining Federated Learning (FL)’s privacy guarantees. This ensures robust and scalable deployment of FL models in dynamic environments, allowing them to generalize effectively despite distribution shifts across decentralized clients.


Recent approaches have explored TTA in FL to enhance model robustness in decentralized environments. FedICON \cite{tan2024} employs contrastive learning to adapt models to diverse client environments, but its high computational demands make it impractical for resource-limited clients. ATP \cite{bao2024} introduces client-specific adaptation by adjusting module-specific adaptation rates. However, it assumes static test-time distributions and does not explicitly encourage inter-client knowledge sharing, which could enhance robustness by leveraging insights from clients with similar data distributions.

Other methods, such as FedTHE+ \cite{jiang2022}, improve personalization and adaptivity by ensembling a global generic classifier and a local personalized classifier in a two-head FL model. However, its performance declines when clients encounter vastly different out-of-distribution data, as combining ensemble classifiers into a more generic global classifier may lead to suboptimal generalization. More recently, FedTSA \cite{zhang2024} introduced a collaboration mechanism using temporal-spatial correlations based on local feature means, allowing clients with similar data distributions to improve personalized model aggregation. However, this method introduces privacy concerns since sharing local feature statistics risks sensitive data leakage through reconstruction. Additionally, FedTSA requires server-side learning (6.2 millions parameters) during adaptation, increasing computational complexity. The reliance on storing local feature means in a memory bank also poses scalability challenges, as memory demands grow with the number of clients and training steps.

To address these limitations, we propose \textbf{Federated Continual Test-Time Adaptation (FedCTTA)}, a framework designed to enable efficient and privacy-preserving adaptation in federated settings. Unlike existing approaches that rely on sharing local feature means in the server side, FedCTTA avoids direct feature sharing, mitigating privacy risks while facilitating continual adaptation. FedCTTA operates by leveraging entropy minimization or updating batch normalization statistics at each client for local test-time adaptation, ensuring model confidence in evolving target distributions. Instead of relying on stored feature statistics, we incorporate similarity-aware aggregation through functional similarity \cite{klabunde2023similarity}, where the server estimates collaborative relationships between participating clients based on client model outputs over a set of randomly generated noise samples. This allows adaptive knowledge sharing while preserving data privacy. Our approach is computationally efficient, as it eliminates the need for additional training on the server side and reduces the memory footprint by avoiding persistent storage of client-specific embeddings. FedCTTA seamlessly integrates continual adaptation with federated learning, ensuring both domain-aware collaboration and robust model generalization without introducing excessive communication or storage overhead. FedCTTA outperforms existing methods under varying degrees of temporal and spatial heterogeneity. It achieves 66.50\% and 63.39\% accuracy under the NIID setting, and 67.78\% and 64.52\% under the IID setting on CIFAR10-C and CIFAR100-C dataset, respectively in TTA-bn method. In contrast, FedTSA achieves 66.19\% and 62.93\% (NIID), as well as 67.51\% and 63.70\% (IID). FedCTTA ensures higher accuracy while preserving privacy in decentralized settings.

Our key contributions are as follows:
\begin{itemize}
    \item We introduce a similarity-aware aggregation technique in federated learning based on functional similarity, where the server calculates collaboration relationships between clients by comparing the outputs of their models over randomly generated noise samples. 
    \item Unlike prior work \cite{jiang2022, zhang2024}, FedCTTA does not store or share local feature embeddings, ensuring data security and mitigating privacy risks.
    \item Our method eliminates the need for server-side training during adaptation, significantly reducing computational complexity.
    \item By avoiding storage of feature means across federated rounds, FedCTTA maintains a constant memory footprint, remaining scalable even as the number of clients or training rounds increases.
\end{itemize}

%% file: sections/2_related_works.tex
\section{Related Works}
\label{sec:related_works}

\subsection{Federated Learning}

Federated learning is a decentralized approach to training machine learning models while keeping data localized, thereby addressing privacy and security concerns. FedAvg \cite{mcmahan2017} aggregates client models into a global one, while FedAvg+FT further fine-tunes it on local data for personalization. FedProx \cite{li2020} introduces regularization to handle client data heterogeneity, and FedAVGM \cite{hsu2019} incorporates momentum for better aggregation.  Li et al. \cite{li2020} proposed using a globally shared dataset to mitigate performance degradation in non-IID data settings, improving model accuracy by up to 30\% on skewed datasets like CIFAR-10. Zhao et al. \cite{zhao2018} introduced FedProx, an extension of FedAvg, to handle statistical and system heterogeneity. FedAMP \cite{huang2021} fosters collaboration between clients with similar data, whereas MOON \cite{li2021} refines local training by leveraging model representation similarity. pFedSD \cite{jin2022} enables clients to distill knowledge from past personalized models, and pFedGraph \cite{ye2023} constructs a collaboration graph based on model similarities. LDAWA \cite{rehman2023} improves aggregation by considering angular divergence, while FedTSA \cite{zhang2024} utilizes temporal-spatial attention to capture both intra-client and inter-client correlations.

\subsection{Test Time Adaptation}

Test-Time Adaptation (TTA) methods enable models to adapt to distribution shifts without access to source data. 
TENT \cite{wang2020} minimizes entropy by updating BatchNorm parameters, achieving state-of-the-art results on corrupted datasets like ImageNet-C with efficient online updates. 
DUA \cite{mirza2022} extends this by dynamically adjusting BatchNorm statistics using minimal unlabeled test data, improving real-time adaptation in scenarios like autonomous driving. 
EATA \cite{niu2022} mitigates catastrophic forgetting and noisy updates through entropy-based sample selection and a Fisher regularizer. 
CoTTA \cite{wang2022} enhances adaptation in non-stationary environments with weight-averaged pseudo-labeling and stochastic restoration of source weights to maintain long-term knowledge. 
These approaches demonstrate diverse strategies for improving TTA efficiency and robustness across various applications.

\subsection{Federated Test Time Adaptation}

Adaptive Test-Time Personalization (ATP) \cite{bao2024} learns module-specific adaptation rates based on client distribution shifts. Clients simulate unsupervised adaptation during training, refining rates to enhance performance on unseen, unlabeled data. FedTHE+ \cite{jiang2022} ensembles global and local classifiers for robust test-time personalization and performs unsupervised fine-tuning, improving accuracy across in-domain (ID) and out-of-domain (OOD) distributions. FedICON \cite{tan2024} uses contrastive learning to capture invariant knowledge from inter-client heterogeneity during training and self-supervision for smooth test-time adaptation, tackling intra-client heterogeneity. While leveraging inter-client heterogeneity to address test-time shifts, FedICON requires extensive contrastive learning, which may be computationally intensive for resource-constrained clients. Xu et al. \cite{xu2023} proposed FedCal, a lightweight framework that performs test-time classifier calibration using estimated label priors from global model predictions. 





%% file: sections/3_methodology.tex
\section{Methodology}
\label{sec:methodology}

\begin{table*}
    \caption{Performance comparison of various federated learning methods with our proposed FedCTTA on CIFAR10-C and CIFAR100-C datasets. We evaluate all methods under two TTA setups: TTA-Grad, where all model parameters are updated during adaptation, and TTA-BN, where only batch normalization layers are updated.}
    \label{tab:all_comp}
    \centering
    \setlength{\tabcolsep}{1.17em}
    \begin{tabular}{lllllllll} 
        \toprule
         & \multicolumn{4}{c}{NIID} & \multicolumn{4}{c}{IID}\\ 
        \cmidrule(lr){2-5}\cmidrule(l){6-9}
         Method & \multicolumn{2}{c}{CIFAR10-C} & \multicolumn{2}{c}{CIFAR100-C} & \multicolumn{2}{c}{CIFAR10-C} & \multicolumn{2}{c}{CIFAR100-C}  \\ 
         \cmidrule(lr){2-3}\cmidrule(lr){4-5}\cmidrule(lr){6-7}\cmidrule(l){8-9}
         & \multicolumn{1}{c}{TTA-grad} & \multicolumn{1}{c}{TTA-bn} & \multicolumn{1}{c}{TTA-grad} & \multicolumn{1}{c}{TTA-bn} & \multicolumn{1}{c}{TTA-grad} & \multicolumn{1}{c}{TTA-bn} & \multicolumn{1}{c}{TTA-grad} & \multicolumn{1}{c}{TTA-bn}\\
        \midrule\midrule
        No-Adapt & 58.47$\pm$0.19 & 58.61$\pm$0.17 & 30.22$\pm$0.12 & 30.22$\pm$0.12 & 58.64$\pm$0.22 & 58.55$\pm$0.21 & 30.22$\pm$0.12 & 30.22$\pm$0.12 \\
        Local    & 63.82$\pm$0.31 & 64.65$\pm$0.29 & 52.85$\pm$0.32 & 55.99$\pm$0.34 & 63.96$\pm$0.33 & 64.79$\pm$0.31 & 52.94$\pm$0.31 & 56.05$\pm$0.34 \\
        FedAvg   & 61.15$\pm$0.24 & 61.45$\pm$0.23 & 51.63$\pm$0.17 & 57.13$\pm$0.43 & 66.12$\pm$0.26 & 67.41$\pm$0.27 & 62.54$\pm$0.31 & 63.96$\pm$0.31 \\
        FedAvg+FT& 63.82$\pm$0.27 & 61.45$\pm$0.23 & 47.83$\pm$0.58 & 57.13$\pm$0.43 & 63.79$\pm$0.30 & 67.41$\pm$0.27 & 61.72$\pm$0.59 & 63.96$\pm$0.31 \\
        FedProx  & 61.68$\pm$0.22 & 61.45$\pm$0.23 & 53.00$\pm$0.38 & 57.13$\pm$0.43 & 66.12$\pm$0.24 & 67.41$\pm$0.27 & 62.33$\pm$0.67 & 63.96$\pm$0.31 \\
        FedAvgM  & 61.50$\pm$0.25 & 61.37$\pm$0.19 & 52.31$\pm$0.46 & 57.13$\pm$0.43 & 63.60$\pm$0.28 & 67.41$\pm$0.27 & 54.66$\pm$0.27 & 63.96$\pm$0.31 \\
        MOON     & 61.58$\pm$0.23 & 61.45$\pm$0.23 & 54.26$\pm$0.27 & 57.13$\pm$0.43 & 66.05$\pm$0.25 & 67.41$\pm$0.27 & 62.40$\pm$0.23 & 63.96$\pm$0.31 \\
        pFedSD   & 61.31$\pm$0.21 & 61.45$\pm$0.23 & 53.33$\pm$0.37 & 57.13$\pm$0.43 & 66.14$\pm$0.26 & 67.41$\pm$0.27 & 62.32$\pm$0.33 & 63.96$\pm$0.31 \\
        pFedGraph& 62.38$\pm$0.26 & 64.21$\pm$0.25 & 57.01$\pm$0.38 & 58.73$\pm$0.38 & 66.10$\pm$0.29 & 64.42$\pm$0.28 & 62.48$\pm$0.30 & 58.75$\pm$0.63 \\
        LDAWA    & 61.85$\pm$0.23 & 61.45$\pm$0.23 & 53.61$\pm$0.33 & 57.13$\pm$0.43 & 65.92$\pm$0.26 & 67.41$\pm$0.27 & 62.37$\pm$0.41 & 63.96$\pm$0.31 \\
        FedTSA   & 63.39$\pm$0.27 & 66.19$\pm$0.26 & 58.03$\pm$0.38 & 62.93$\pm$0.29 & 66.29$\pm$0.28 & 67.51$\pm$0.27 & 62.62$\pm$0.36 & 63.70$\pm$0.34 \\
        \mrule
        \rowcolor[gray]{0.9} \ours & \textbf{66.23$\pm$0.28} & \textbf{66.50$\pm$0.27} & \textbf{64.81$\pm$0.29} & \textbf{63.39$\pm$0.28} & \textbf{66.64$\pm$0.29} & \textbf{67.78$\pm$0.28} & \textbf{64.15$\pm$0.28} & \textbf{64.52$\pm$0.28} \\
        \brule
    \end{tabular}
\end{table*}

\subsection{Problem Definition}
Continual test-time adaptation (TTA) addresses the challenge of adapting models to sequentially arriving, non-stationary target domains without access to source data. In federated settings, where clients observe distinct or overlapping domains that evolve over time, this becomes even more challenging. We consider a federated system with $N$ clients, $\mathcal{C} = \{C_1, C_2, \dots, C_N\}$, each receiving a data stream $\mathcal{D}_t^{(i)}$ over time. The objective is to adapt the client models $\theta$ while preserving privacy and preventing catastrophic forgetting, ensuring robust performance despite evolving data distributions.

\subsection{Local Test-time Adaptation: Entropy Minimization}

In the client side, we explore two approaches for continual Test-Time Adaptation (TTA): 
(1) Fine-tuning all model parameters via entropy minimization, and 
(2) Updating only the batch normalization (BN) layer statistics (mean and variance).

In the first approach, entropy minimization is employed to align the feature distributions with evolving target domains. For an input \( x \in \mathcal{D}_t^{(i)} \), the entropy \( H(p) \) is defined as:

\begin{equation}
    H(p) = -\sum_{k=1}^K p_k \log(p_k),
\end{equation}

where \( p = f_{\theta_i}(x) \) is the predicted probability vector, and \( K \) is the number of classes. The entropy minimization objective for client \( C_i \) is given by:

\begin{equation}
    L_{\text{ent}} = \frac{1}{|\mathcal{D}_t^{(i)}|} \sum_{x \in \mathcal{D}_t^{(i)}} H(f_{\theta_i}(x)).
\end{equation}

The model parameters \( \theta \) are updated to minimize \( L_{\text{ent}} \) via gradient descent. Minimizing \( L_{\text{ent}} \) encourages confident predictions (low uncertainty), facilitating feature alignment with the target domain.

In the second approach, only the running mean \( \mu \), variance \( \sigma^2 \), and scale/shift parameters \( \gamma \) and \( \beta \) of the BN layers are updated for each incoming data stream:

\begin{equation}
    \mu_i^{\text{new}} = (1 - \alpha) \mu_i^{\text{old}} + \alpha \cdot \mathbb{E}_{x \sim \mathcal{D}_t^{(i)}}[x],
\end{equation}

\begin{equation}
    \sigma_i^{2, \text{new}} = (1 - \alpha) \sigma_i^{2, \text{old}} + \alpha \cdot \text{Var}(x \sim \mathcal{D}_t^{(i)}),
\end{equation}

where \( \alpha \) is the momentum parameter, \( \mathbb{E}[\cdot] \) represents the batch mean, and \( \text{Var}[\cdot] \) denotes the batch variance.

By combining BN statistics updates and full parameter updates through entropy minimization, both methods enable domain-specific adaptation. This allows each client to locally adjust its model to the test domain, improving accuracy, mitigating catastrophic forgetting, and ensuring privacy in federated settings.

\subsection{Similarity-Aware Aggregation}
To facilitate efficient collaboration, model aggregation at the server leverages functional similarity \cite{klabunde2023similarity}, defined as the similarity in output behavior or probability distributions across clients. Clients exposed to similar domains contribute more substantially to each other’s model updates while maintaining data privacy. As direct data sharing between clients is prohibited, the server generates a reference dataset composed of random noise samples, denoted as \( D_{\text{noise}} = \{ z_i \}_{i=1}^M \). These noise samples are evaluated by each client model to produce corresponding output logits. The server then computes pairwise client similarities based on these logits using various distance metrics, including cosine similarity, cross-entropy, and negative Euclidean distance. Empirical results demonstrate that negative Euclidean distance yields the best performance within this framework, as reported in Table \ref{tab:comparison}.


For client \( i \), the logits for a random noise sample \( z \) are represented as \( f_{\theta_i}(z) \). The similarity between clients \( i \) and \( j \) is computed using the negative Euclidean distance between their mean logits:

\begin{equation}
    D_{ij} = -\| \mu_i - \mu_j \|_2,
\end{equation}

where \( \mu_i \) and \( \mu_j \) are the mean logits over the random noise dataset:

\begin{equation}
    \mu_i = \frac{1}{M} \sum_{k=1}^M f_{\theta_i}(z_k), \quad 
    \mu_j = \frac{1}{M} \sum_{k=1}^M f_{\theta_j}(z_k).
\end{equation}

A higher (less negative) \( D_{ij} \) indicates greater similarity between clients. To derive collaboration weights, the pairwise distances \( D_{ij} \) are normalized using the softmax function  \( C_{ij} \) and the server performs weighted aggregation based on \( C_{ij} \), and each client updates its local model accordingly:

\begin{equation}
    \theta_i^{\text{new}} = \sum_{j=1}^K \frac{\exp(D_{ij})}{\sum_{k=1}^K \exp(D_{ik})} \theta_j,
\end{equation}

where \( C_{ij} \) represents the contribution of client \( j \) to client \( i \)'s aggregation, and \( K \) is the total number of clients. \( \theta_j \) denotes the parameters of client \( j \)'s model. This approach fosters domain-aware collaboration by prioritizing updates from similar clients, improving adaptation and continuous learning in federated settings.

\begin{table*}[ht]
    \caption{Detailed performance comparison under spatial IID and temporal heterogeneity using the TTA-bn method.}
    \label{tab:spa_iid}
    \centering
    \setlength{\tabcolsep}{0.63em}
    \begin{tabular}{l|c|ccccccccccccccc|c}
        \trule
        & Time & \multicolumn{15}{c|}{t \makebox[0.75\textwidth]{\rightarrowfill}} &  \\
        \mrule
        \vcell{\rotv{Datasets}} & \vcell{Method} & \rot{Gaussian} & \rot{Shot} & \rot{Impulse} & \rot{Defocus} & \rot{Glass} & \rot{Motion} & \rot{Zoom} & \rot{Snow} & \rot{Frost} & \rot{Fog} & \rot{Brightness} & \rot{Contrast} & \rot{Elastic} & \rot{Pixelate} & \rot{Jpeg} & \rot{Mean} \\[-\rowheight]
        \pcb & \pcm & \pcb & \pcb & \pcb & \pcb & \pcb & \pcb & \pcb & \pcb & \pcb & \pcb & \pcb & \pcb & \pcb & \pcb & \pcb & \pcb \\         
        \dmrule
        \multirow{7}{*}{\rotv{CIFAR10-C}} & Source & 37.30 & 38.44 & 26.08 & 28.99 & 33.92 & 27.44 & 30.21 & 34.53 & 32.89 & 10.63 & 36.46 & 23.53 & 37.51 & 41.43 & 43.70 & 30.54 \\
        & Local     & 69.76 & 69.81 & 63.21 & 68.69 & 61.40 & 63.92 & 66.58 & 67.14 & 67.44 & 55.35 & 71.16 & 39.68 & 64.93 & 70.25 & 71.58 & 64.72 \\
        & FedAvg    & 72.28 & 72.46 & 66.31 & 71.08 & 64.22 & 66.45 & 69.61 & 70.03 & 69.34 & 58.32 & 73.60 & 42.84 & 68.34 & 72.92 & 74.55 & 67.49 \\
        & pFedGraph & 68.40 & 69.45 & 62.88 & 68.35 & 61.48 & 63.82 & 65.72 & 66.70 & 66.57 & 55.56 & 70.37 & 40.52 & 65.14 & 69.28 & 70.96 & 64.34 \\
        & FedTSA    & 72.56 & 72.72 & 66.06 & 71.13 & 64.25 & 66.35 & 69.39 & 70.10 & 69.66 & 58.74 & 73.61 & 42.61 & 68.47 & 72.98 & 74.69 & 67.55 \\
        \cmrule{2-18}
        & \gcell\ours & \gcell73.21 & \gcell73.04 & \gcell66.27 & \gcell71.91 & \gcell64.64 & \gcell66.32 & \gcell69.58 & \gcell70.24 & \gcell70.38 & \gcell57.57 & \gcell74.20 & \gcell41.93 & \gcell68.16 & \gcell73.72 & \gcell75.00 & \gcell67.74 \\
        
        \dmrule
        \multirow{7}{*}{\rotv{CIFAR100-C}} & Source & 14.05 & 16.64 & 34.76 & 41.60 & 19.35 & 38.15 & 43.32 & 36.48 & 27.63 & 20.96 & 54.91 & 17.24 & 35.02 & 11.45 & 41.73 & 30.22 \\
        & Local & 51.59 & 53.02 & 50.26 & 65.13 & 50.77 & 63.23 & 65.07 & 58.10 & 58.17 & 51.10 & 66.70 & 61.25 & 56.67 & 59.98 & 51.57 & 57.51 \\
        & FedAvg & 57.33 & 58.60 & 56.74 & 69.07 & 57.51 & 68.94 & 70.92 & 64.29 & 64.19 & 57.44 & 72.40 & 67.36 & 63.70 & 66.33 & 57.56 & 63.50 \\
        & pFedGraph & 52.05 & 53.42 & 50.48 & 65.60 & 51.19 & 63.61 & 65.49 & 58.39 & 58.64 & 51.39 & 67.08 & 61.64 & 57.14 & 60.46 & 52.06 & 57.91 \\
        & FedTSA & 57.56 & 58.75 & 57.23 & 69.73 & 56.27 & 69.18 & 71.05 & 64.33 & 64.60 & 56.44 & 73.10 & 67.77 & 63.30 & 66.58 & 58.21 & 63.61 \\
        \cmrule{2-18}
        & \gcell\ours & \gcell56.90 & \gcell59.58 & \gcell57.06 & \gcell72.12 & \gcell58.47 & \gcell70.04 & \gcell71.84 & \gcell65.32 & \gcell65.66 & \gcell58.27 & \gcell74.34 & \gcell68.41 & \gcell64.29 & \gcell67.32 & \gcell58.97 & \gcell64.57 \\
        \brule   
    \end{tabular}
\end{table*}

%% file: sections/4_experimental_results.tex
\section{Experimental Results}
\label{sec:results}

\subsection{Implementation Details}
\subsubsection{Datasets and Setup}

We evaluate our proposed aggregation method on two standard corruption benchmarks: CIFAR10-C and CIFAR100-C, where a model trained on CIFAR-10 and CIFAR-100, respectively, is adapted to their corrupted versions. These datasets are constructed by applying 15 distinct corruption types at five severity levels to the test and validation images of the original CIFAR datasets. Consistent with prior work, we report the average accuracy across all corruption types and clients at the highest severity level (severity 5). For CIFAR-100 to CIFAR100-C adaptation, we utilize a pretrained ResNeXt-29 model obtained from the Robustbench benchmark\cite{croce2021}, while for CIFAR10-C, we employ a pretrained ResNet-8 \cite{He2015}. To simulate a dynamically evolving test distribution, we progressively alter the corruption type at severity 5 over time. The test data is distributed among 20 clients to emulate a federated learning (FL) setting with decentralized data. Each client processes streaming test data in batches of 10, experiencing sequences of distribution shifts.

Test-Time Adaptation (TTA) methods aim to improve the robustness of a pretrained model when handling unlabeled test data. One approach, TTA-bn, adjusts batch normalization (BN) statistics to match the test distribution, as demonstrated by NORM \cite{Schneider2020} and DUA \cite{mirza2022}, without requiring gradient computation. In contrast, TTA-grad methods \cite{wang2020, wang2022} adapt the model using backpropagation with self-supervised losses. Our proposed method focuses on fostering inter-client collaboration to share knowledge across clients and is orthogonal to the TTA strategies employed by individual clients. We evaluate our approach under both TTA-bn and TTA-grad settings for each dataset. For TTA-grad, we adopt entropy minimization as the local adaptation strategy, optimizing with SGD optimizer using a learning rate of \(1.0 \times 10^{-5}\).

\subsubsection{Baselines}
We compare our proposed method with FedAvg and other regularization-based FL methods, including \textbf{FedAvg+FT}, \textbf{FedProx}, \textbf{FedAvgM}, \textbf{MOON}, and \textbf{pFedSD}, which we have adapted for test-time adaptation. Additionally, we evaluate our proposed method alongside other personalized federated learning (PFL) and test-time adaptation (TTA) methods.
\textbf{pFedGraph} \cite{ye2023} constructs a collaboration graph based on model similarities and dataset size to enhance collaboration at the server side. \textbf{LDAWA} \cite{rehman2023} aggregates model weights by measuring angular divergence between a client's model and the global model and adjusting the aggregation accordingly.
\textbf{FedTSA} \cite{zhang2024} leverages a temporal-spatial attention module to capture both intra-client temporal correlations and inter-client spatial correlations. 

\subsubsection{Spatial and Temporal Heterogeneity}
To quantify heterogeneity in our experimental setup, we adopt the notions of temporal heterogeneity and spatial heterogeneity as defined in \cite{zhang2024}. These metrics characterize the distribution shifts encountered by clients during continual test-time adaptation.

 \textbf{Spatial Heterogeneity: }Spatial heterogeneity at time \( t \), denoted as \( SH_t \), measures the diversity of data distributions among clients:
    \begin{equation}
    SH_t = \frac{N_{\text{cls}}}{N}
    \end{equation}
where \( N_{\text{cls}} \) is the number of client clusters with consistent distribution shifts, and \( N \) is the total number of clients. Higher \( SH_t \) values indicate greater heterogeneity, with \( SH_t = 1 \) signifying unique distribution shifts for all clients.

\textbf{Temporal Heterogeity: } Temporal heterogeneity for the \( i \)-th client, denoted as \( TH_i \), measures the frequency of distribution changes in streaming data:
    \begin{equation}
    TH_i = \frac{T_{\text{con}}}{T}
    \end{equation}
where \( T_{\text{con}} \) is the total duration of time slots with consistent distribution shifts, and \( T \) is the total duration of all time slots. Higher \( TH_i \) values indicate greater heterogeneity, with \( TH_i = 1 \) signifying a distinct distribution shift in every time slice.

\begin{table*}
    \caption{The experimental setup and performance in the NIID scenario, where Clients 1–4, Clients 5–7, and Clients 8–10 share similar data distributions throughout the entire lifecycle, with a total of 10 clients.}
    \label{tab:diif_client}
    \centering
    \setlength{\tabcolsep}{0.693em}
    \begin{tabular}{c|ccccccccccccccc|c}
        \trule
        Time & \multicolumn{15}{c|}{t \makebox[0.77\textwidth]{\rightarrowfill}} &  \\
        \mrule
        \vcell{Client 1-4} & \rot{Gaussian} & \rot{Shot} & \rot{Impulse} & \rot{Defocus} & \rot{Glass} & \rot{Motion} & \rot{Zoom} & \rot{Snow} & \rot{Frost} & \rot{Fog} & \rot{Brightness} & \rot{Contrast} & \rot{Elastic} & \rot{Pixelate} & \rot{Jpeg} & \rot{Mean} \\[-\rowheight]
        \pcm & \pcb & \pcb & \pcb & \pcb & \pcb & \pcb & \pcb & \pcb & \pcb & \pcb & \pcb & \pcb & \pcb & \pcb & \pcb & \pcb \\ 
        \mrule
        \vcell{Client 5-7} & \rot{Glass} & \rot{Motion} & \rot{Zoom} & \rot{Snow} & \rot{Frost} & \rot{Fog} & \rot{Brightness} & \rot{Contrast} & \rot{Elastic} & \rot{Pixelate} & \rot{Jpeg} & \rot{Gaussian} & \rot{Shot} & \rot{Impulse} & \rot{Defocus} & \rot{Mean} \\[-\rowheight]
        \pcm & \pcb & \pcb & \pcb & \pcb & \pcb & \pcb & \pcb & \pcb & \pcb & \pcb & \pcb & \pcb & \pcb & \pcb & \pcb & \pcb \\ 
        \mrule
        \vcell{Client 8-10} & \rot{Jpeg} & \rot{Pixelate} & \rot{Elastic} & \rot{Contrast} & \rot{Brightness} & \rot{Fog} & \rot{Frost} & \rot{Snow} & \rot{Zoom} & \rot{Motion} & \rot{Glass} & \rot{Defocus} & \rot{Impulse} & \rot{Shot} & \rot{Gaussian} & \rot{Mean} \\[-\rowheight]
        \pcm & \pcb & \pcb & \pcb & \pcb & \pcb & \pcb & \pcb & \pcb & \pcb & \pcb & \pcb & \pcb & \pcb & \pcb & \pcb & \pcb \\ 
        \dmrule
        No-Adapt & 71.56 & 66.60 & 58.06 & 46.70 & 59.88 & 47.00 & 60.02 & 63.48 & 51.10 & 48.36 & 51.22 & 43.92 & 64.44 & 71.44 & 71.68 & 58.36 \\
        pFedGraph & 68.54 & 66.06 & 61.72 & 54.42 & 66.40 & 57.04 & 67.60 & 54.36 & 65.54 & 61.98 & 69.18 & 50.86 & 65.46 & 67.56 & 67.66 & 62.95\\
        FedTSA & 68.70 & 67.30 & 63.02 & 56.58 & 66.48 & 57.36 & 68.32 & 57.26 & 66.38 & 62.08 & 69.18 & 52.72 & 66.58 & 68.20 & 69.76 & 63.99\\
        \mrule
        \rowcolor[gray]{0.9} \ours & 68.64 & 67.84 & 67.20 & 58.66 & 67.76 & 58.56 & 70.52 & 58.98 & 67.18 & 64.00 & 70.26 & 57.02 & 67.68 & 70.56 & 70.22 & 65.67\\
        \brule     
    \end{tabular}
\end{table*}

\subsection{Performance Analysis}

We assessed the performance of our method in both TTA-grad and TTA-bn settings under two distinct scenarios, with data distributed across 20 clients. In the first scenario, we simulated spatial heterogeneity ($SH_t$ = 0.2) with 4 clusters, which we refer to as NIID, while the second scenario involved very low spatial heterogeneity ($SH_t$ = 0.05) with a single cluster, referred to as IID. For both scenarios, temporal heterogeneity ($TH_i$) was kept constant at 0.02. Since TTA-bn does not require backward optimization, unlike many state-of-the-art methods such as FedProx, MOON, pFedSD, and LDAWA that rely on gradient updates, the performance of these methods under TTA-bn is consistent with that of FedAvg. 

As shown in Table \ref{tab:all_comp}, the baseline (No-adapt) struggles with corrupted datasets, achieving low accuracy across all settings, underscoring the challenges of distribution shifts in federated learning, particularly under non-IID scenarios. Local adaptation strategies improve performance in IID settings but remain inadequate against severe shifts. While FedAvg and other regularization-based methods such as FedProx, FedAvgM, pFedSD, and MOON perform well in IID settings, their effectiveness declines in NIID scenarios, where personalized federated learning (PFL) methods like FedTSA and pFedGraph achieve better results.

FedTSA and pFedGraph show notable improvements over previous methods, particularly under the NIID setting. They outperform FedAvg and FedProx in both CIFAR10-C and CIFAR100-C, highlighting the importance of inter-client collaboration for improving performance under corrupted conditions. Our method outperforms all other approaches across all settings in both CIFAR10-C and CIFAR100-C, achieving the highest accuracy. This demonstrates the effectiveness of our method in handling both spatial and temporal distribution shifts in federated learning scenarios.


\begin{figure}
    \centering
    \begin{subfigure}{0.49\columnwidth}
        \centering
        \includegraphics[width=\linewidth]{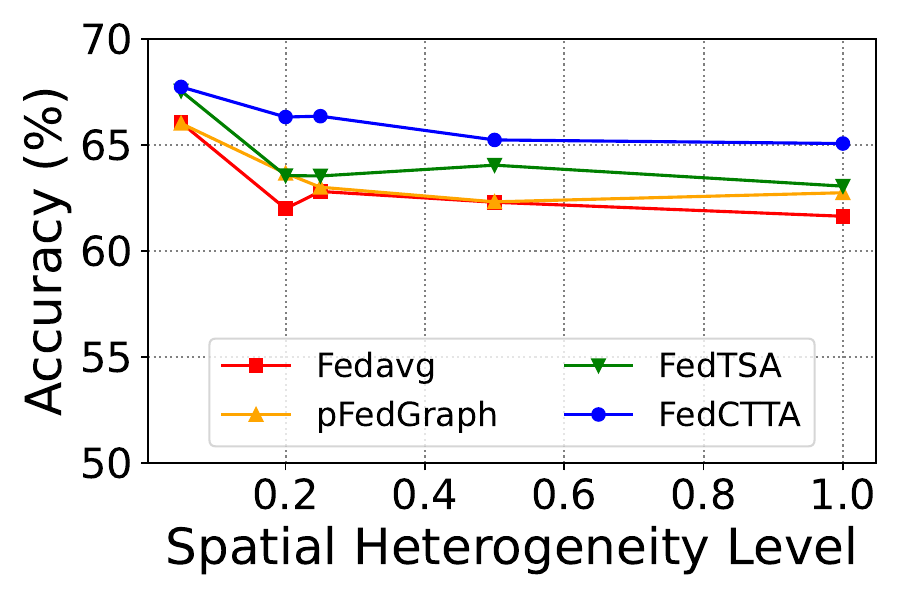}
        \caption{CIFAR10-C}
        \label{fig:cifar10_spatial}
    \end{subfigure}
    \hfill
    \begin{subfigure}{0.49\columnwidth}
        \centering
        \includegraphics[width=\linewidth]{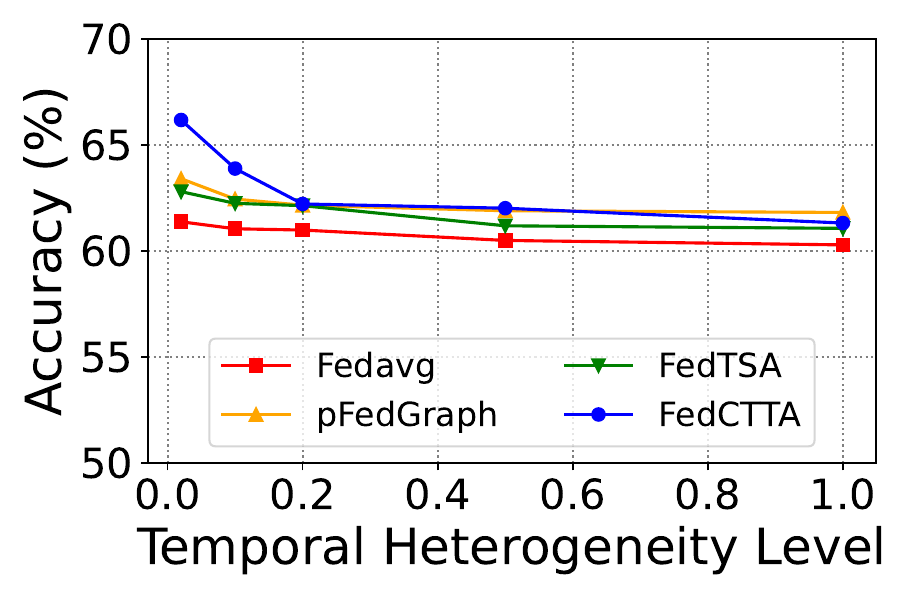}
        \caption{CIFAR10-C}
        \label{fig:cifar10_temporal}
    \end{subfigure}
    
    \begin{subfigure}{0.49\columnwidth}
        \centering
        \includegraphics[width=\linewidth]{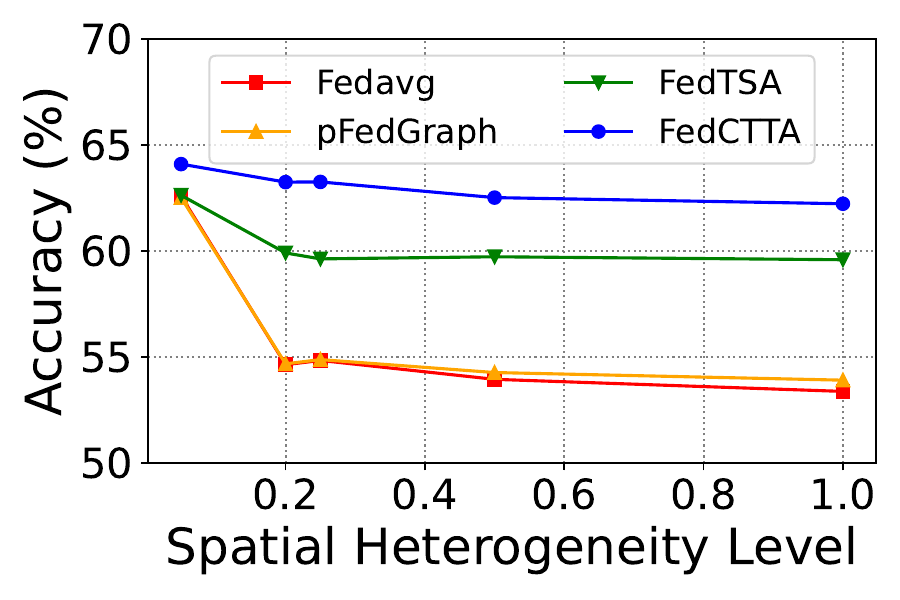}
        \caption{CIFAR100-C}
        \label{fig:cifar100_spatial}
    \end{subfigure}
    \hfill
    \begin{subfigure}{0.49\columnwidth}
        \centering
        \includegraphics[width=\linewidth]{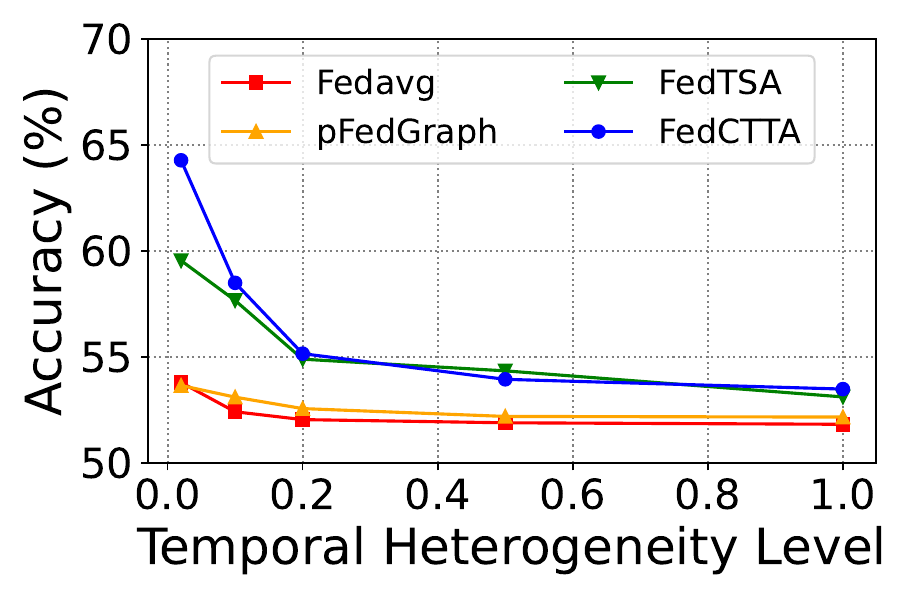}
        \caption{CIFAR100-C}
        \label{fig:cifar100_temporal}

    \end{subfigure}

    \caption{Comparison of accuracy on CIFAR-10C and CIFAR-100C under varying degrees of temporal and spatial heterogeneity.}
    \label{fig:comparison}
    \vspace{-5mm}
\end{figure}

\begin{figure}[!ht]
    \centering
    \begin{subfigure}{\columnwidth}
        \centering
        \includegraphics[width=0.9\columnwidth]{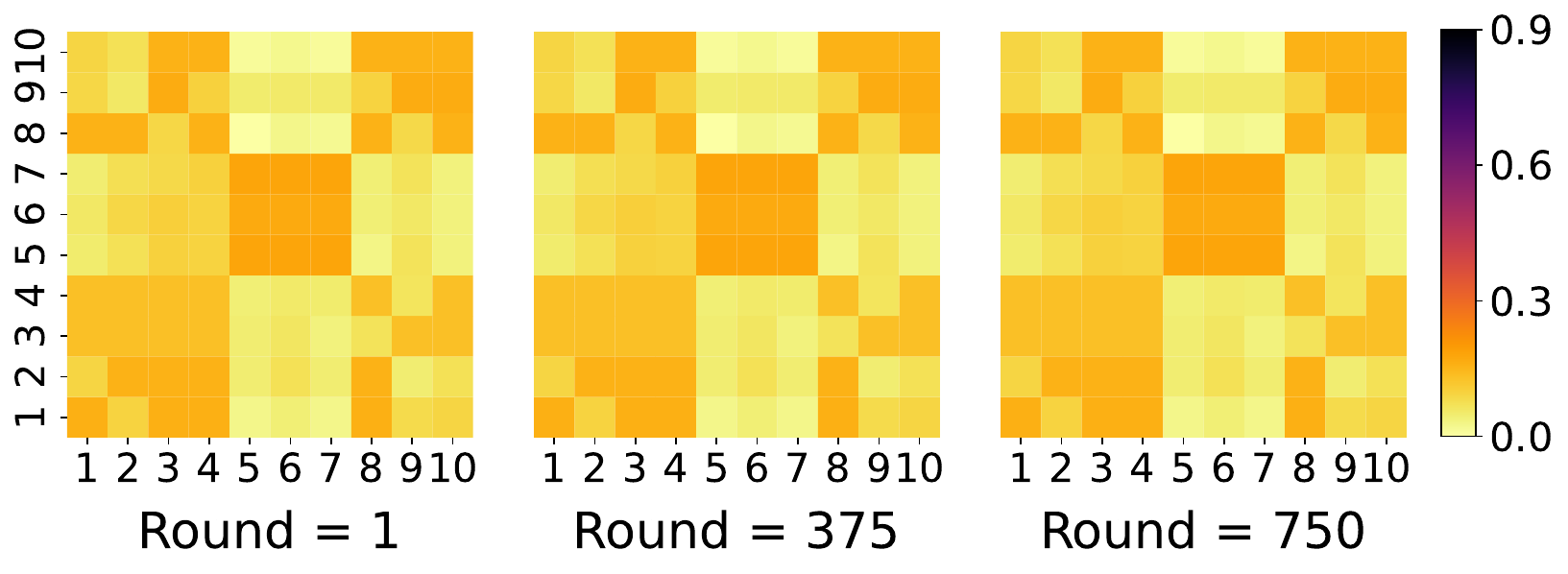}
        \caption{pFedGraph}
        \label{fig:collab_fedgraph}
    \end{subfigure}
    
    \begin{subfigure}{\columnwidth}
        \centering
        \includegraphics[width=0.9\columnwidth]{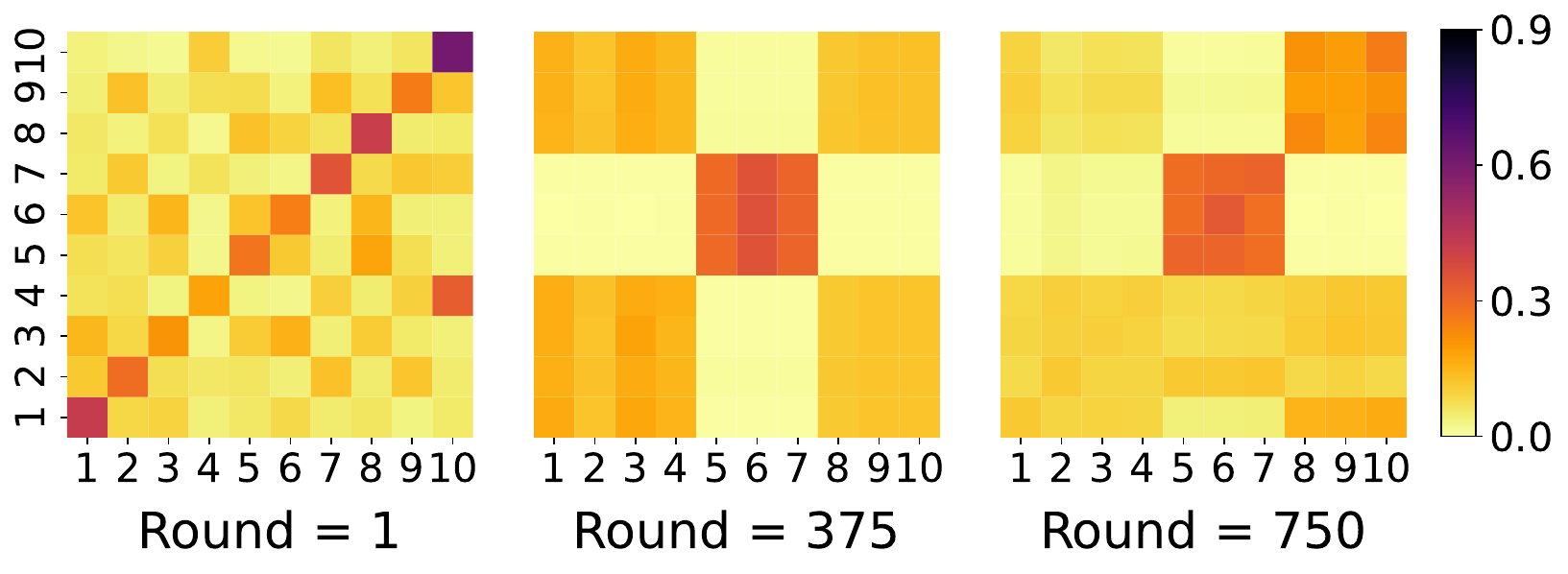}
        \caption{FedTSA}
        \label{fig:collab_fedtsa}
    \end{subfigure}
    
    \begin{subfigure}{\columnwidth}
        \centering
        \includegraphics[width=0.9\columnwidth]{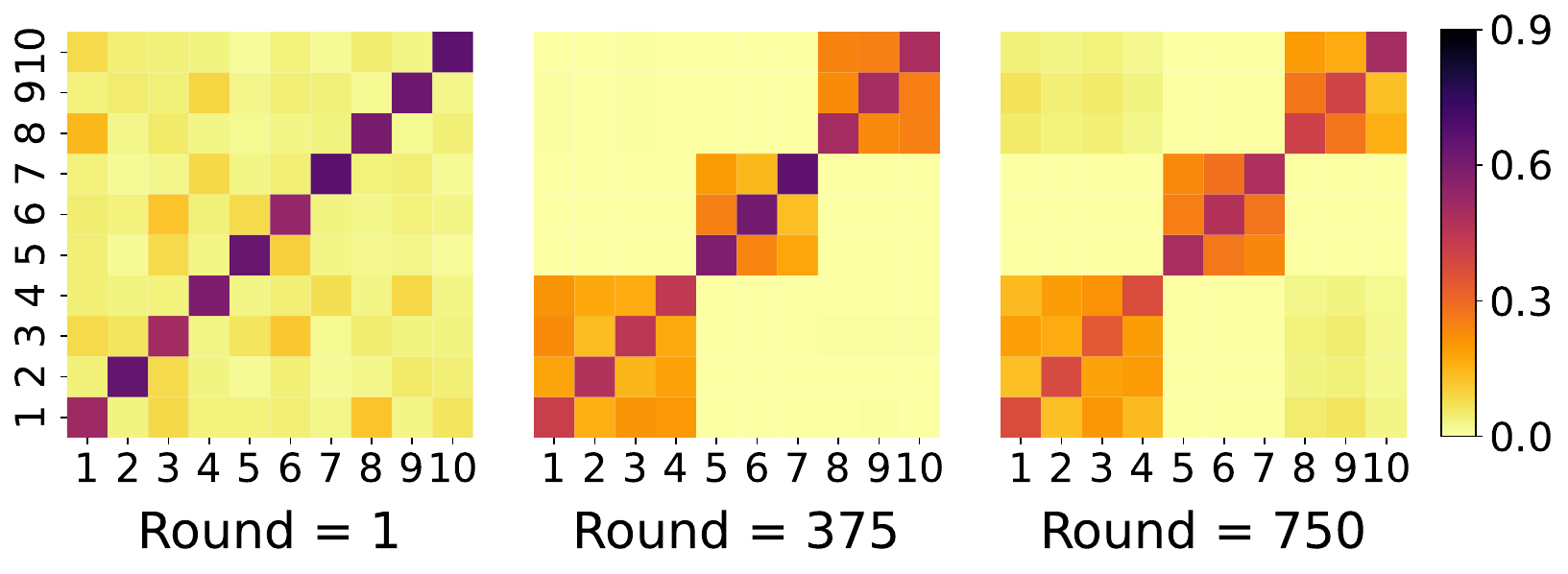}
        \caption{\ours}
        \label{fig:collab_fedctta}
    \end{subfigure}
    
    \caption{The evolution of the collaboration matrix across federated rounds for three methods: pFedGraph, FedTSA, and FedCTTA. In FedCTTA, clients with similar data distributions naturally form clusters in aggregation weighting.}
    \label{fig:collab_matrix}
    \vspace{-5mm}
\end{figure}

\subsection{Robustness Under Spatial and Temporal Heterogenity}

To assess the robustness of our proposed method, we conduct experiments on CIFAR10-C and CIFAR100-C under varying degrees of spatial and temporal heterogeneity. Specifically, when analyzing spatial heterogeneity (SH), we fix temporal heterogeneity ($TH_i$) at 0.02, and conversely, when varying $TH_i$, we maintain $SH_t$ at 0.2 to isolate the impact of each factor independently.

As shown in Figure \ref{fig:cifar10_spatial} and \ref{fig:cifar100_spatial}, our method consistently outperforms the baselines across different levels of spatial heterogeneity. While all methods experience a decline in accuracy as heterogeneity increases, FedAvg suffers the most significant drop, indicating its poor adaptability to spatially non-iid data. Our method demonstrates strong resilience, with only a minor performance decline, highlighting its ability to effectively handle diverse client distributions. 

As illustrated in Figure \ref{fig:cifar10_temporal} and \ref{fig:cifar100_temporal}, our method also exhibits strong robustness against temporal heterogeneity, outperforming traditional and personalized federated learning baselines in most cases. The core strength of our approach lies in its adaptive aggregation strategy, which leverages temporal similarity between clients to facilitate more effective inter-client collaboration. When temporal heterogeneity is low, distribution shifts occur gradually, allowing our method to retain and utilize historical knowledge more effectively. However, as temporal heterogeneity increases, abrupt shifts in data distribution diminish the relevance of past information. In case of high temporal heterogeneity ($TH_i$ = 1), our method performs comparably to FedTSA.

\begin{figure}[!h]
    \centering
    \includegraphics[width=0.9\linewidth, keepaspectratio]{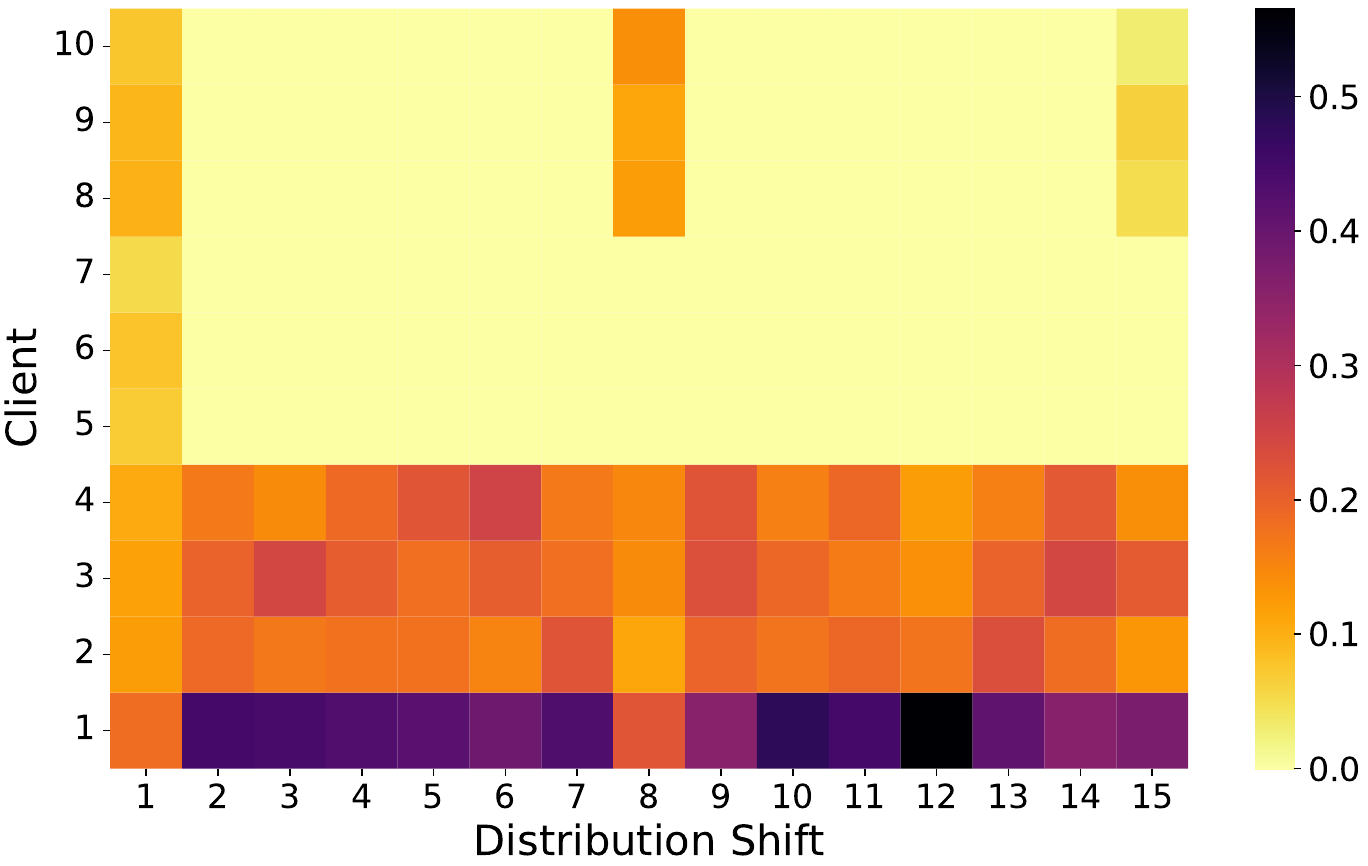}
    \caption{Time-varying collaboration matrix for Client 1 of our proposed FedCTTA method under NIID setting. Throughout all rounds, Clients 1–4 observe the same data distribution. At 8th distribution shift, Clients 8–10 also observe data from the same domain as Clients 1–4, and therefore have higher similarity with Client 1.}
    \label{fig:collab_client1}
    \vspace{-2mm}
\end{figure}

\subsection{Collaboration Relationship Analysis}

Figure \ref{fig:collab_matrix} illustrates the evolution of the collaboration matrix across federated rounds for three methods: pFedGraph, FedTSA, and our proposed FedCTTA. The collaboration matrix quantifies the aggregation weights between clients, where higher values indicate stronger collaboration. This case study evaluates 10 clients on CIFAR10-C, divided into three groups based on the sequence of distribution shifts: Group 1 (Clients 1-4), Group 2 (Clients 5-7), and Group 3 (Clients 8-10). In Figure \ref{fig:collab_fedgraph}, pFedGraph exhibits a scattered collaboration pattern across federated rounds, lacking structured inter-client relationships. Figure \ref{fig:collab_fedtsa} shows that FedTSA initially relies on self-updates, with limited collaboration emerging over time, but without well-defined client clusters. In contrast, Figure \ref{fig:collab_fedctta} demonstrates that FedCTTA naturally clusters clients with similar data distributions, fostering structured and adaptive collaboration. Figure \ref{fig:collab_client1} further illustrates the time-varying collaboration matrix for Client 1 in this setup. Throughout all rounds, Clients 1–4 observe the same data distribution, forming a distinct cluster. At the 8th distribution shift, Clients 8–10 begin observing data from the same domain as Clients 1–4, leading to an increase in similarity with Client 1. These results highlight FedCTTA’s effectiveness in leveraging inter-client similarities, where collaboration is determined based on similarity between output logits evaluated on random noise samples. A detailed quantitative comparison under different distribution shifts is provided in Table \ref{tab:diif_client}.

%% file: sections/5_ablation_study.tex
\section{Ablation Study}
\label{sec:ablation}

\begin{figure}[!ht]
    \centering
    \begin{subfigure}{\columnwidth}
        \centering
        \includegraphics[width=0.9\linewidth, keepaspectratio]{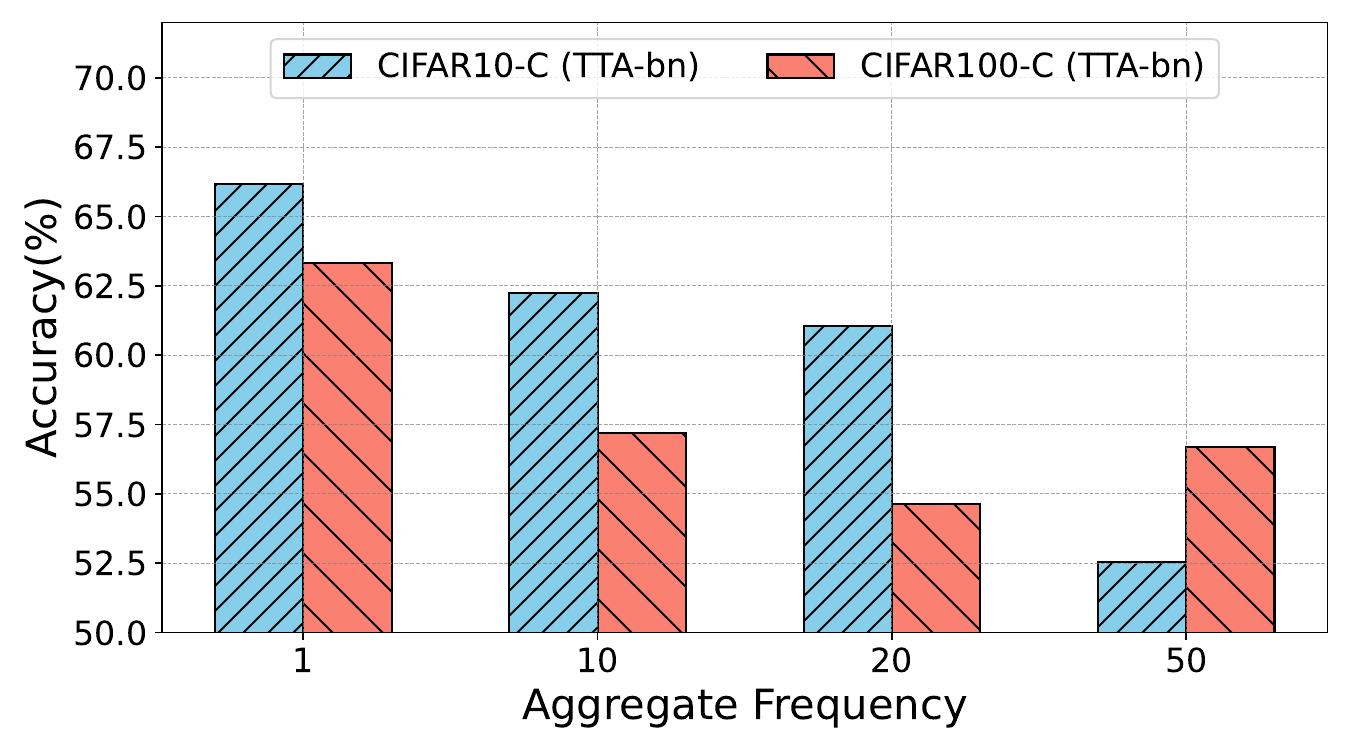}
        \caption{Effect of aggregation frequency}
        \label{fig:agg_freq}
    \end{subfigure}
    
    \begin{subfigure}{\columnwidth}
        \centering
        \includegraphics[width=0.9\linewidth, keepaspectratio]{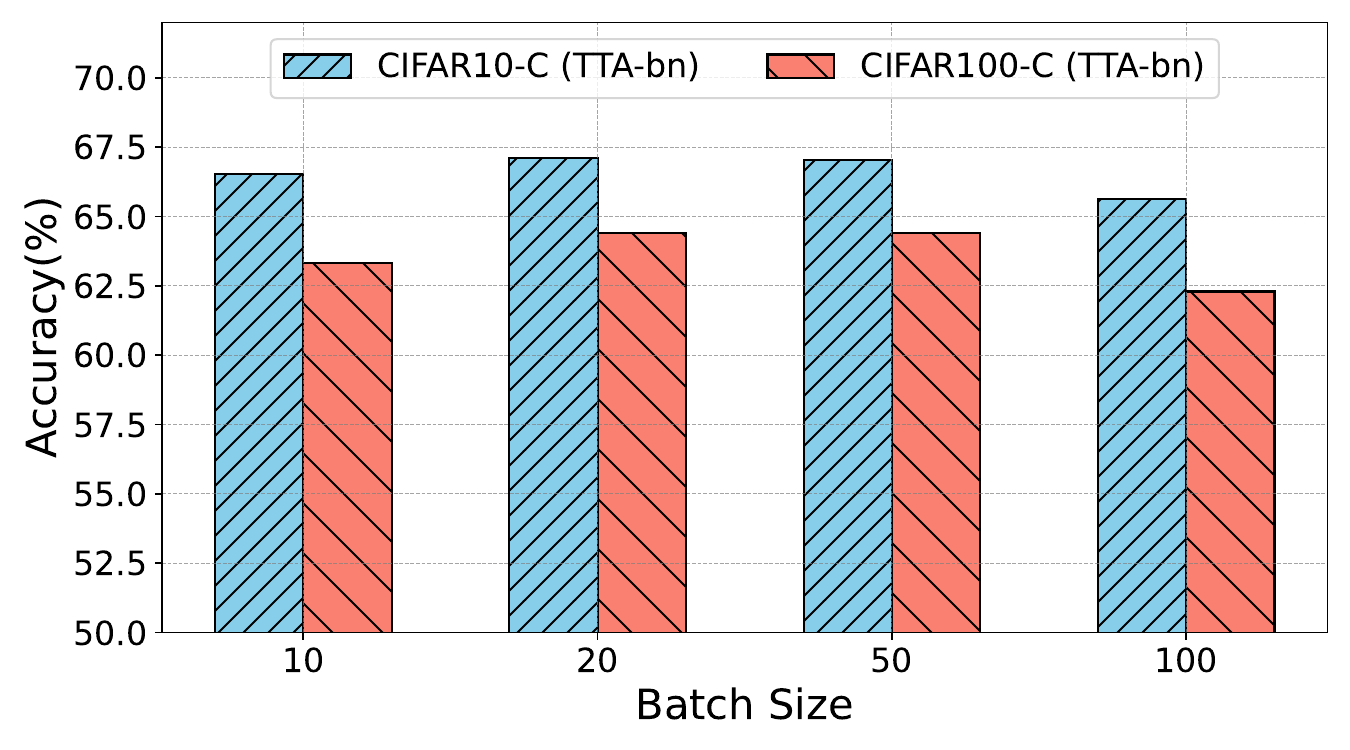}
        \caption{Effect of batch size}
        \label{fig:batch_size}
    \end{subfigure}

    \caption{(a) Gradual performance decline observed as aggregation interval increases across federated rounds. (b) Both very high and very low batch sizes impact performance, affecting generalization and stability. A balanced batch size is ideal.}
    \label{fig:ablation}
    \vspace{-5mm}
\end{figure}

\subsection{Effect of Aggregation Frequency}
We analyze the impact of aggregation frequency on test accuracy for CIFAR10-C and CIFAR100-C using the TTA-bn method. As shown in Figure \ref{fig:agg_freq}, increasing the aggregation interval negatively affects performance across federated rounds. A higher aggregation interval (e.g., 50) leads to reduced accuracy, suggesting that frequent model updates and collaboration between clients are crucial for maintaining performance. 

\subsection{Effect of Batch Size}
Figure \ref{fig:batch_size} illustrates the effect of batch size on accuracy. We observe that both very low (10) and very high (100) batch sizes result in suboptimal performance. A moderate batch size (20 or 50) achieves better results. In the federated setup, each client receives a smaller number of samples per domain, and when the batch size is too large, more frequent distribution shifts occur, leading to reduced performance. Conversely, a very small batch size can cause unstable updates, impacting accuracy. This trend is consistent across both datasets.

\begin{table}[ht]
    \caption{Comparison of test accuracy using distance measures for output logits and feature embeddings on CIFAR10-C dataset with the TTA-grad method under the NIID setting.}
    \label{tab:comparison}
    \centering
    \setlength{\tabcolsep}{0.55em}
    \begin{tabular}{ccccccc}
        \toprule
        \multirow{2}{*}{Data} & \multicolumn{4}{c}{Output Logit (Acc. \%)} & \multicolumn{2}{c}{Feature (Acc. \%)} \\
        \cmidrule(lr){2-5} \cmidrule(lr){6-7}
        & Euclid & KL-div & CE & Cosine & Euclid & Cosine \\
        \midrule
        Random Noise & 66.19 & 61.62 & 61.60 & 61.62 & 62.07 & 61.63 \\
        Selected (CIFAR) & 65.92 & 61.65 & 61.64 & 61.63 & 61.80 & 61.63 \\
        \bottomrule
    \end{tabular}
\end{table}

\subsection{Ablation on Distance Metric and Auxiliary Dataset}
Table \ref{tab:comparison} compares test accuracy using different distance measures for output logits and feature spaces, evaluated on random noise and selected CIFAR10-C samples with the TTA-grad method under the NIID setting. The results show that deriving output logits from random noise and using negative Euclidean distance to build the collaboration matrix for personalized model aggregation achieves the best performance.

%% file: sections/6_conclusion.tex
\section{Conclusion}
\label{sec:conclusion}
We propose FedCTTA, a privacy-preserving and efficient framework for continual test-time adaptation (CTTA) in federated learning (FL). FedCTTA enables adaptive knowledge transfer across evolving data distributions using similarity-aware aggregation without sharing feature embeddings and integrates entropy minimization for confident adaptation. Experiments on CIFAR10-C and CIFAR100-C demonstrate that FedCTTA outperforms existing methods in accuracy, robustness, and scalability under spatial and temporal heterogeneity, while maintaining low computational and memory overhead. Future work will explore integrating advanced CTTA techniques and testing on more dynamic real-world datasets.